\def\year{2022}\relax
\documentclass[letterpaper]{article} 
\usepackage{aaai22}  
\usepackage{times}  
\usepackage{helvet}  
\usepackage{courier}  
\usepackage[hyphens]{url}  
\usepackage{graphicx} 
\usepackage{natbib}  
\usepackage{caption} 
\DeclareCaptionStyle{ruled}{labelfont=normalfont,labelsep=colon,strut=off} 
\usepackage{amsmath}
\usepackage{amsfonts}
\usepackage{amsthm}
\usepackage{amssymb}

\usepackage[ruled, linesnumbered, vlined]{algorithm2e}

\usepackage{url}
\usepackage{subfigure}
\usepackage{mathrsfs} 
\usepackage{hhline} 
\usepackage{makecell}
\usepackage{booktabs}
\usepackage{multirow}
\usepackage{soul}
\setulcolor{red} 
\setstcolor{blue} 
\sethlcolor{yellow} 

\usepackage{graphicx}
\usepackage{appendix}

\usepackage{textcomp}

\usepackage{makecell}
\usepackage{threeparttable}

\usepackage{diagbox}

\title{iGrow: A Smart Agriculture Solution to Autonomous Greenhouse Control}
\author{
    Xiaoyan Cao\equalcontrib \textsuperscript{\rm 1},
    Yao Yao\equalcontrib \textsuperscript{\rm 2},
    Lanqing Li\textsuperscript{\rm 3},
    Wanpeng Zhang\textsuperscript{\rm 2},
    Zhicheng An\textsuperscript{\rm 2},
    Zhong Zhang\textsuperscript{\rm 3},
    Li Xiao\textsuperscript{\rm 2},
    Shihui Guo\textsuperscript{\rm 1},
    Xiaoyu Cao\textsuperscript{\rm 4},
    {Meihong Wu}\footnote{Corresponding authors.}\textsuperscript{\rm 1},
    {Dijun Luo}\footnotemark[2]\textsuperscript{\rm 3}
}

\affiliations {
    \textsuperscript{\rm 1}School of Informatics, Xiamen University,
    \textsuperscript{\rm 2}Tsinghua Shenzhen International Graduate School, Tsinghua University,
    \textsuperscript{\rm 3}Tencent AI Lab,
    \textsuperscript{\rm 4}College of Chemistry and Chemical Engineering, Xiamen University
    
    \textsuperscript{\rm 1}\{holmescao@stu., guoshihui@, wmh@\}xmu.edu.cn,
    \textsuperscript{\rm 2}\{y-yao19@mails, zhangwp19@mails, azc19@mails, xiaoli@sz\}.tsinghua.edu.cn,
    \textsuperscript{\rm 3}\{lanqingli, todzhang, dijunluo\}@tencent.com,
    \textsuperscript{\rm 4}xcao@xmu.edu.cn
    
}

\begin{document}

\maketitle

\begin{abstract}
Agriculture is the foundation of human civilization. However, the rapid increase of the global population poses a challenge on this cornerstone by demanding more food. Modern autonomous greenhouses, equipped with sensors and actuators, provide a promising solution to the problem by empowering precise control for high-efficient food production. However, the optimal control of autonomous greenhouses is challenging, requiring decision-making based on high-dimensional sensory data, and the scaling of production is limited by the scarcity of labor capable of handling this task. With the advances of artificial intelligence (AI), the internet of things (IoT), and cloud computing technologies, we are hopeful to provide a solution to automate and smarten greenhouse control to address the above challenges. In this paper, we propose a smart agriculture solution named iGrow, for autonomous greenhouse control (AGC): (1) for the first time, we formulate the AGC problem as a Markov decision process (MDP) optimization problem; (2) we design a neural network-based simulator incorporated with the incremental mechanism to simulate the complete planting process of an autonomous greenhouse, which provides a testbed for the optimization of control strategies; (3) we propose a closed-loop bi-level optimization algorithm, which can dynamically re-optimize the greenhouse control strategy with newly observed data during real-world production. We not only conduct simulation experiments but also deploy iGrow in real scenarios, and experimental results demonstrate the effectiveness and superiority of iGrow in autonomous greenhouse simulation and optimal control. Particularly, compelling results from the tomato pilot project in real autonomous greenhouses show that our solution significantly increases crop yield (+10.15\%) and net profit (+92.70\%) with statistical significance compared to planting experts. Our solution opens up a new avenue for greenhouse production. The data and code are provided in supplementary materials.
\end{abstract}

\section{Introduction}

With the global challenge in food caused by the continuous population growth, the transformation and upgrading of the agricultural industry is urgently in need
~\cite{deichmann2016will}. 

As a result, the greenhouse industry is expanding rapidly due to the ability to provide fresh food steadily~\cite{leeallen2015accelerated}.
This ability is attributed to its controlled indoor environment, which provides a favorable environment for crop growth.
In particular, modern high-tech autonomous greenhouses integrated with IoT (sensors and actuators) and cloud computing technologies support real-time remote monitoring and precise control, thus promising high crop yields at relatively low resource costs.
However, autonomous greenhouse management nowadays relies mainly on experienced labor and the scarcity of such human resources limits the production scale-up~\cite{brain15current}. 
On the other hand, when facing a high volume of multimodal information (such as long time-series of temperature, humidity, CO$_2$ concentration, etc.), it is not feasible to rely solely on human decisions to control high-tech autonomous greenhouses.

With the development of AI, new data-driven technologies are being applied in various agriculture subfields~\cite{liakos2018machine}.
To the best of our knowledge, there is no mature AI-based solution for the autonomous greenhouse control (AGC) optimization problem.
The main task of the AGC optimization problem is to provide a control strategy that aims to maximize crop yield and minimize resource consumption of a long-term planting period (typically lasting 3-5 months).
However, data collection in greenhouse production is costly and time-consuming, accompanied by sparse signals (e.g., fruits weight). From a machine learning point of view, this poses a serious challenge of insufficient samples for any data-driven optimization algorithms.
Inspired by the idea of digital twins, we consider a simulator to simulate the autonomous greenhouse planting process (abbreviated as AGPP), which enables fast generation of virtual planting trajectories.
With a simulator as a testbed, the AGC problem becomes a long-horizon strategic optimization problem, similar to robot control~\cite{vemula2016path}. Then one can use optimization algorithms (e.g., reinforcement learning (RL), heuristic algorithms) to explore the optimal AGC strategy on the simulator.

In the literature of greenhouse modeling, the dynamic simulation of the greenhouse planting process is mainly divided into indoor climate simulation~\cite{van2010optimal} and crop growth simulation~\cite{marcelis2008simulating}.
Traditional rule-based simulators rely on strong assumptions and constraints according to expert knowledge (such as crop growth mechanisms), which may oversimplify the planting process~\cite{van2010optimal,marcelis2008simulating}.
Although this type of simulator reduces the reliance on data, its limited expressive power can lead to a gap between simulation and reality.
As a universal approximator~\cite{hornik1989multilayer}, the powerful ability of NN-based models in the representation of complex nonlinear systems has attracted more and more attention~\cite{Chellapilla1999Evolving, C2008Artificial}.

Several studies develop NN-based simulators to simulate the climate or crop yields in greenhouses~\cite{schillaci2021prediction,salazar2014tomato}.
However, these simulators do not couple indoor climate and crop growth states together to simulate the complete planting process in autonomous greenhouses.
In addition, they consider only a few factors affecting the greenhouse planting process, for example, in yield prediction, only CO$_2$ concentration, transpiration and solar radiation intensity are used as inputs~\cite{salazar2014tomato}.
To be a competent testbed for strategy optimization, the complete the AGPP simulator should comprehensively consider the interaction of greenhouse climate, crop growth state, and greenhouse control strategy. To this end, we propose an NN-based simulator, which involves 14 factors (such as temperature, humidity, CO$_2$ concentration, illumination, etc.) to approximate the complete AGPP as much as possible.

The quantity and quality of data are key to training an accurate enough NN-based simulator.
IoT technologies enable measurement of high dimensional data~\cite{mekala2017survey}, which can be efficiently processed and stored via the cloud platform~\cite{keerthi2015cloud}. The combination of IoT and the cloud platform provides an ideal solution for data collection and management. However, limited by the sensor capacity, only partial observations of the greenhouse states can be monitored. Any NN-based simulator trained on such collected data inevitably leads to the simulation-to-reality gap.
To alleviate this gap, we introduce an incremental scheme in our framework. The intuition behind this is to incrementally update the simulator by continuously accumulating sensor data, thus keep improving the accuracy and can be eventually qualified for deployment to large-scale real-world production.

In this paper, we formalize the AGC problem and propose a smart agriculture solution named iGrow, which is built upon AI, IoT and cloud-native technologies. An overview of iGrow is presented in Figure~\ref{fig:framework}.
The main contributions of our work are summarized as follows:

\begin{figure*}[thbp]
    \centering
    \includegraphics[width=0.8\linewidth,height=0.25\linewidth]{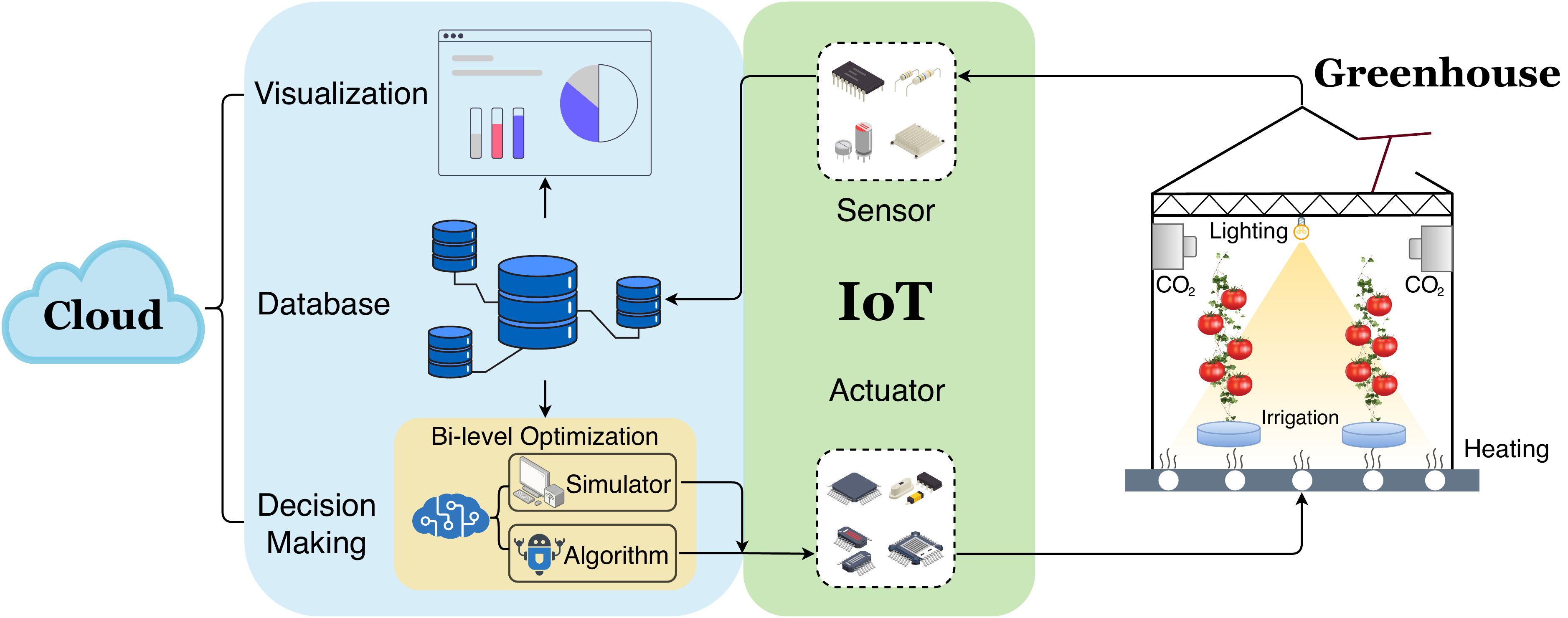}
    \caption{An overview of iGrow.
    Specifically, iGrow optimizes the AGC strategy based on a decision-making module, then performs actions and monitors greenhouse status through IoT technology. Sensor data is visualized on the cloud platform, as well as used by a bi-level optimization algorithm to dynamically re-optimize the decision-making module to calibrate errors.}
    \label{fig:framework}
\end{figure*}

\begin{itemize}
    \item We formulate the AGC problem as a Markov decision process (MDP) optimization problem.

    \item We are the first to propose an incremental NN-based three-stage simulator to continuously approximate the dynamic of the complete AGPP.
    The proposed simulator is validated on both virtual and real tomato datasets with accuracy comparable to the state-of-the-art (SOTA) rule-based simulator.
    
    \item We propose a closed-loop bi-level optimization algorithm to dynamically iterate control strategies during the AGPP. At the lower-level, we use incremental real planting data to calibrate the simulator; at the upper-level, we use a heuristic algorithm to re-optimize control strategies based on the latest simulator.  
    
    \item We test iGrow in both simulated and real greenhouses scenarios (take tomato as an example), and experimental results demonstrate the technical and economic value of iGrow in autonomous greenhouse simulation and optimal control.
    Particularly, during a tomato pilot project in real autonomous greenhouses, experimental results show the performance of iGrow statistically significantly ($Sig < 0.01$) exceeds that of planting experts with an average improvement of 10.15\% in yield and 92.70\% in net profit.
    
\end{itemize}

\section{Related Work}

\subsection{Greenhouse planting process simulation}
The dynamic simulation of the greenhouse planting process can be divided into two parts: modeling of the underlying (1) physical dynamics (climate simulator) and (2) biological process (crop simulator)~\cite{marcelis2008simulating, van2010optimal}. 
There is a set of research focusing on the design of mechanistic physical dynamics equations to simulate the dynamic greenhouse climate~\cite{dincer2001energy, pinon2005constrained, van2010optimal}. 
As the representation power of NN for complex tasks is widely verified, some methods such as recurrent network~\cite{fitz2011neural} and long short-term memory network~\cite{schillaci2021prediction} have been applied to simulate greenhouse climate change.

Crop modeling is the other essential part of greenhouse planting process simulation, which can be used to predict yield, growth, etc~\cite{marcelis2008simulating}. 
Horticultural researchers focus on exploring crop growth mechanisms and establish some rule-based crop models~\cite{bertin1993dry, marcelis2008simulating}.
However, crop growth is a complex nonlinear system, thus the potential relationships between variables are difficult to be characterized.
To improve the predictability of crop models, NN-based simulators have been explored on crop yield prediction, such as sweet peppers~\cite{lin2008neural}, tomato~\cite{salazar2014tomato, an2021simulator}.

To the best of our knowledge, AI methods for modeling the complete AGPP remain underexplored. In this paper, we present an incremental NN-based simulator that combines indoor climate, crop growth state, and environmental control operations to simulate the AGPP.

\subsection{Optimal control for greenhouse}

IoT with cloud computing technology has been used to provide smart agricultural solutions~\cite{mekala2017survey}. Some applications~\cite{van2015optimal, parameswaran2016arduino} utilize wireless sensor networks to monitor and control crop growth. A smart irrigation system has been developed for remote control to minimize human involvement~\cite{parameswaran2016arduino}. In order to improve wheat yields, a reinforcement learning algorithm is used to optimize soil variables on a wheat yield prediction model~\cite{garcia1999use}. For crop planting in autonomous greenhouses, researchers use the model predictive control algorithm to adjust the actual state of greenhouse climate close to target state~\cite{van2015optimal, hemming2020cherry, zhang2021robust}. 

However, there are no proven solutions to provide automatic strategies for greenhouse control. 
In this paper, we propose a closed-loop bi-level optimization algorithm to dynamically iterate control strategies during the AGPP.

\section{Problem Statement}
\label{sec:problem_statement}
In this work, we formulate the AGC problem as a stochastic MDP optimization problem. In the following section, we give the necessary background of MDP and formally introduce the AGC optimization problem.

\subsection{MDP introduction}
An MDP can be denoted by a tuple $\mathbb{S}=\left(\mathcal{S}, \mathcal{A}, \mathcal{P}, \mathcal{R}, \gamma,\rho_0 \right)$, where $\mathcal{S} \subseteq \mathbb{R}^{n}$ and $\mathcal{A} \subseteq \mathbb{R}^{m}$ represent the state and action spaces, respectively. The dynamics or transition distribution are denoted as $\mathcal{P}\left(s^{\prime} \mid s,a\right)$, the initial state $s_0 \in \mathcal{S}$ is assumed to follow the initial state distribution $\rho_0$, $\mathcal{R}(s,a,s^{\prime})$ gives the reward function of executing action $a$ to transition from state $s$ to $s^{\prime}$, and $\gamma \in(0,1]$ is the discount factor. A policy $\pi(a \mid s)$ on $\mathbb{S}$ maps a state $s$ to a probability distribution over $\mathcal{A}$ that generates trajectories as $\tau=\left(s_{0}, a_{0}, r_{0}, s_{1}, a_{1}, \ldots, s_{T}\right)$, where $a_{t} \sim \pi \left( \cdot\mid s_{t}\right)$, $s_{t+1} \sim \mathcal{P}\left(\cdot \mid s_{t}, a_{t}\right)$, $r_{t}=\mathcal{R}\left(s_{t}, a_{t}, s_{t+1}\right)$, and $T$ is the terminate time step. 

\subsubsection{MDP optimization problem.}  \emph{Find a policy $\pi^*$ that maximizes the cumulative expected return $R_T$ given $s_0$:}
\begin{equation}
    \pi^{*}=\underset{\pi}{\operatorname{argmax}} E_{\tau}\left[R_{T}\right]=\underset{\pi}{\operatorname{argmax}} E_{\tau}\left[\sum_{t=0}^{T-1} \gamma^{t} r_{t}\right].
\end{equation}


\subsection{Formulation}
\label{sesc:AGC}
An complete AGPP can be denoted by a tuple $\mathscr{G}=<C, G, Y, A, W>$, where $c \in C \subseteq \mathbb{R}^4$ denotes greenhouse climate, $g \in G \subseteq \mathbb{R}^3$ is the crop growth state, $y \in Y \subseteq \mathbb{R}^{+} \cup \{0\}$ represents the yield, $a \in A \subseteq \mathbb{R}^4$ specifies the control setpoints, and $w \in W \subseteq \mathbb{R}^6$ represents the outside weather.

The objective of AGC is to find a strategy that both improves crop yields and reduces expenses, such as resource consumption and labor costs, and this can be viewed as a specific instance of the MDP optimization problem. There are many factors involved in the AGC optimization problem, but we can only obtain partial observations from the greenhouse due to the limitation of the sensors. As a result, we focus on 14 observable factors that have a significant impact on planting in the autonomous greenhouse, and omit other variables. Within this context, we can specify the state space, the action space, the reward function, and the transition function of the AGC problem as follows. 
\begin{itemize}

  \item State space: In the AGC problem, each state $s$ consists of a four tuple $<w, c, g, y>$, and details of each element are shown in Supplementary Table S1. It is noteworthy that each variable of the state is a number with different units, which encode the different information determining the growth status of the crop. 
  
  \item Action space: Only four essential control variables are involved in the AGC problem, including temperature, CO$_2$ concentration, illumination, and irrigation. We present basic statistics in Supplementary Table S2. Notice that the setting of these action variables will have a decisive effect on the transition of state variables, except for the outside weather $w$, which is beyond control.

  \item Reward function: Since AGC aims to weigh crop yields against total expenses, we denote the cumulative return $R_T$ by the crop gains minus the cost of control strategy consisting of resource consumption, labor costs, etc. By setting $\gamma=1$, the reward function $\mathcal{R}$ can be converted into the formulation:  $r_t=R_{t+1}-R_{t}$.

  \item Transition function: The transition function in AGC problem is assumed to be unknown, and we seek for a simulator of the dynamics $\hat{\mathcal{P}}\left(s^{\prime} \mid s,a\right)$ as the approximation of $\mathcal{P}\left(s^{\prime} \mid s,a\right)$.

\end{itemize}

\subsection{Bi-level optimization}
Let $\theta$ be the parameterization of the simulator $\Hat{\mathcal{P}}$, and parameterize the control strategy $\pi$ by parameters $\phi$. In AGC problem, $\Hat{\mathcal{P}}_\theta$ should be learned before the optimization of strategy $\pi_{\phi}$ in consideration of the sample inefficiency of crop planting. Within this context, the AGC optimization can be formulated as:
\begin{equation}
\label{equ:bi-level}
    \begin{array}{c}
    \max _{\phi} R_{t}\left(\theta^{*}, \phi\right) \\
    \text { s.t. } \theta^{*}=\text{argmin}_{\theta} L_{\text {train }}(\theta; D),
    \end{array}
\end{equation}
where $\mathcal{L}_{train}(\theta; D)$ is the training loss on a given dataset $D$. This formulation is consistent with bi-level optimization~\cite{wen1991linear} in a broader scope since both $\phi$ and $\theta$ need to be optimized to achieve better strategies. $\phi$ and $\theta$ are treated respectively as upper-level and lower-level variables that are optimized in an interleaving way.
In particular, the lower-level optimization represents the simulator calibration with continuous data collection, while the upper-level optimization corresponds to the strategy iterations on the calibrated simulator.

\section{Methodology}

\label{sec:Methodology}

As shown in Figure~\ref{fig:framework}, we lay out the framework designed for the AGC optimization problem. The proposed solution consists of several essential components and each of them plays a different role in bi-level optimization. To be specific, the decision-making module leverages optimization algorithms to optimize the control strategy on a given simulator, which essentially is a predictive model learned from greenhouse planting data in our context. Supported by IoT and cloud technologies, the optimized strategy can be deployed to real greenhouses, and the corresponding greenhouse state will be monitored and stored in the database. We recall that the continuous collection of new planting data will undoubtedly contribute to the calibration of the data-driven NN-based simulator, which provides a more accurate testbed for further optimization of the AGC control strategy.

\subsection{Incremental NN-based simulator}
\label{sec:Incremental_model}

As mentioned above, the simulator $\Hat{\mathcal{P}}$ of complete the AGPP is a prerequisite to solving the AGC optimization problem.
Since data-driven methods can simulate the planting dynamic with less prior knowledge, we use NNs to build the simulator for the sake of high accuracy and generalizability. 
Inspired by the design pattern of rule-based simulators, our proposed simulator is divided into three modules ($\mathbb{C}_{\Theta_{1}}$, $\mathbb{G}_{\Theta_{2}}$ and $\mathbb{Y}_{\Theta_{3}}$) instead of directly using one-module modeling to approximate $\mathcal{P}$.
The three-module design is equivalent to introducing an additional regularization that not only enhances the interpretability of the simulator but also prevents overfitting while reducing the demand for training data.
The detailed modeling process is described below and the configuration of the three neural networks is given in Technical Appendix Section 2.1.

\subsubsection{Greenhouse climate module:} 

\begin{figure}
    \centering
    \includegraphics[width=0.48\textwidth]{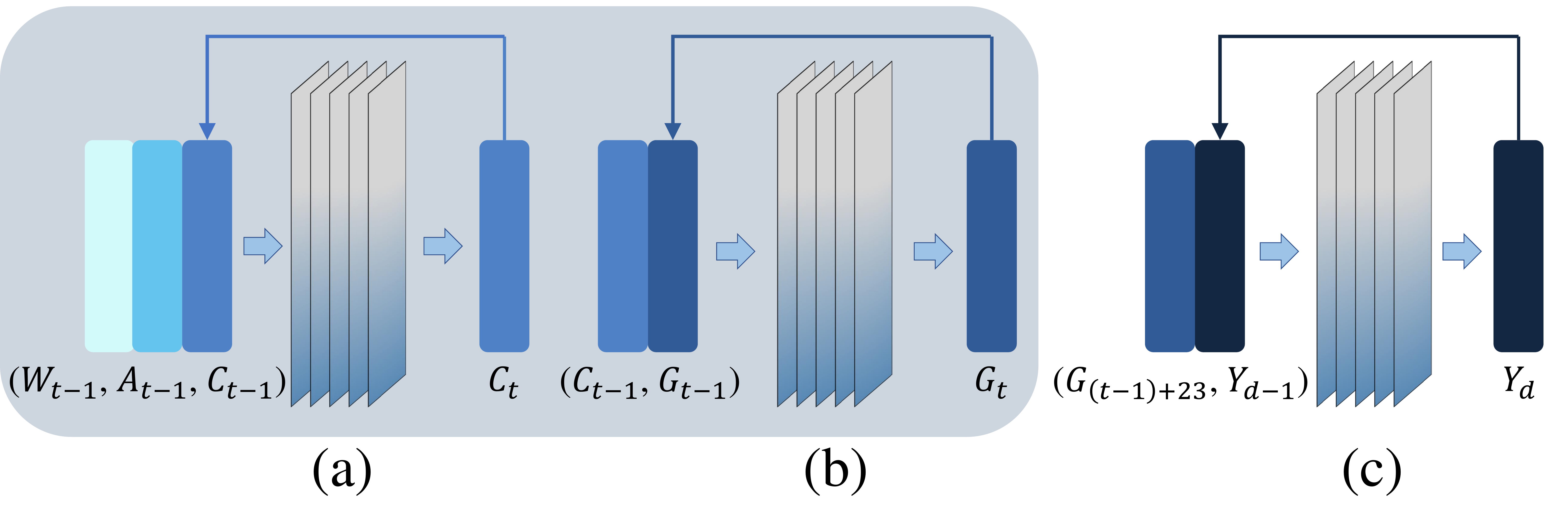}
    \caption{Network structure of the incremental simulator. (a) and (b) represent the greenhouse climate module and the crop growth module, respectively, which simulate the greenhouse planting process at hourly level. (c) represents the yield module, which simulates at daily level.}
    \label{fig:Incremental_model}
\end{figure}

The indoor climate $c_{t}$ at time $t$ can be predicted by the previous outside weather $w_{t-1}$, control setpoints $a_{t-1}$, and indoor climate $c_{t-1}$ according to a specific greenhouse climate simulator, denoted by $\mathbb{C}$. As shown in Figure~\ref{fig:Incremental_model}, we propose an NN-based structure, parameterized by $\Theta_{1}$, to simulate the transition function of indoor climate change: $W\times A \times C \rightarrow C$, where $W$, $A$ and $C$ are defined in Section \ref{sesc:AGC}.
We adopt the mean-square error (MSE) as the loss function.
Moreover, we assume that greenhouse climate states transit once per hour, which is a suitable time granularity to approximate reality.

\subsubsection{Growth state module:} 

It is known from horticultural experience and biological knowledge that crop growth is mainly influenced by indoor climate during the planting process~\cite{marcelis2008simulating}.
Hence, we build the growth state module $\mathbb{G}_{\Theta_{2}}$ to approximate the transition function of crop growth: $C\times G \rightarrow G$, where $\Theta_{2}$ represents the corresponding neural network parameters.
Note that the loss function and the time granularity of this module is the same as the greenhouse climate module. 

\subsubsection{Yield module:} 

Different from greenhouse climate and growth state, the change in crop yield within one hour is negligible, and we focus on daily yield estimation in this work. According to literature~\cite{bertin1993dry}, the crop yield is determined by crop growth state. Therefore, the process of yield accumulation process can be formulated by $G\times Y \rightarrow Y$.
In order to solve the problem of inconsistent time granularity with the growth state module, we extend $G$ to a vector $\vec{G}_d=\left[g^{(0)}_d,\ldots, g^{(23)}_d\right]$, where $g^{(i)}_d$ represents the growth state of the $i$-th hour of day $d$, so that the yield module can be denoted by: $y_{d} \leftarrow {\mathbb{Y}_{\Theta_{3}}}(g_{d-1}^{\left(23\right)}, y_{d-1})$. Similarly, we use the MSE loss function to train the parameters $\Theta_{3}$ of the yield module.

\subsubsection{Incremental mechanism:} 

Note that there is an inherent simulation-to-reality gap between $\hat{\mathcal{P}}$ and $\mathcal{P}$ due to partial observations and limited data.
However, according to the central limit theorem~\cite{rosenblatt1956central}, we can deduce that $\lim\limits_{|D| \to +\infty }\hat{\mathcal{P}} = \mathcal{P}$, 
where $|D|$ represents the scale of the dataset. 
This inference manifests that given sufficiently abundant real data, the data-driven simulator $\hat{\mathcal{P}}$  can be an ideal approximation of the dynamic $P$ in real greenhouses.
Therefore, we introduce the concept of the incremental mechanism based on the three-stage simulator, that is, streaming updates the simulator with the newly collected data for error calibration.

\subsection{Optimization algorithms}
\label{sec:Optimization algorithmsl}
AI optimization algorithms are known to be an efficient way to solve the MDP optimization problem~\cite{kolobov2012planning, yao2021sample}.
With the incremental simulator as the testbed, we consider two classical AI optimization algorithms, including the elitist genetic algorithm (EGA)~\cite{rani2019effectiveness} and the soft actor-critic algorithm (SAC)~\cite{haarnoja2018soft}, to seek the optimal control strategy for the AGC problem.
The detailed description and procedure of them are given in Technical Appendix Section 1.

\subsection{AGC bi-level optimization}
\label{sec:control_procedure_iGrow}

When there is a gap between the simulator and the real planting process, the gap has a bad impact on the simulator-based algorithm optimization.
That is, there maybe larger bias between the simulated strategy performance and that of in the real deployment~\cite{janner2019trust}.
To alleviate this problem, we propose a bi-level optimization algorithm to achieve continuous but possibly asynchronous optimization of the simulator and the strategy, as shown in Algorithm~\ref{algo:Decision_making}. 
In a complete autonomous greenhouse planting period $T$, the control strategy is re-optimized on the latest simulator at every $K_1$ time step.
The setpoints generated by the latest strategy are fed back to the greenhouse to adjust the state of the greenhouse.
Then, the setpoints and the state of the greenhouse will be collected in the data buffer.
On the other hand, for every $K_2$ time step, the simulator will in turn be calibrated by fine-tuning on the latest data buffer.
The proposed procedure is similar to lifelong learning~\cite{field2000lifelong} and allows the flexibility to incrementally iterate the simulator and the greenhouse control strategy at different paces to ensure stable and optimal progress. 

\begin{algorithm}[t]
\DontPrintSemicolon
\KwIn{Dataset $D$, simulator $\hat{\mathcal{P}}$, the update period $K_1$ of strategy, and the update period $K_2$ of simulator}
\KwOut{Updated simulator $\hat{\mathcal{P}}$ and updated dataset $D$}
Initialize $\tau = \varnothing $\\
\tcp{A complete planting period $T$}
\For{$t \in [0, T) $}{
\tcp{The upper-level optimization}
\If(){$t \ mod \ K_1 == 0$}{
$\pi \leftarrow$ re-optimize $\pi$ by simulating on the latest simulator $\hat{\mathcal{P}}$\\ 
}
\tcp{The lower-level optimization}
\If(){$t \ mod \ K_2 == 0$}{
$\hat{\mathcal{P}} \leftarrow$ update $\hat{\mathcal{P}}$ by $\tau$\\
}
\tcp{Control and monitoring greenhouse status}
$a_t \leftarrow \pi$, $s_{t+1} \leftarrow \mathcal{P}(s_t, a_t), r_{t} \leftarrow \mathcal{R}(s_t, a_t)$\\
\tcp{Cumulative planting data}
$\tau \leftarrow \tau \cup (s_t, a_t, r_{t}, s_{t+1})$ \\
}
$D \leftarrow D \cup {\tau}$
\caption{AGC bi-level optimization algorithm}
\label{algo:Decision_making}
\end{algorithm}

\begin{table}[t]
\begin{tabular}{@{}c|c|c|c|c@{}}
\hline
\diagbox{Sim}{Var}      & AirT            & AirRH           & AirCO$_2$          & PAR              \\ \cline{1-5}
Baseline    & 0.915           & 0.707           & 0.930              & 1.000            \\ \hline
Incremental & 0.986           & 0.941           & 0.988              & 1.000            \\ \hline\hline
\diagbox{Sim}{Var}      & LAI             & PlantLoad       & NetGrowth          & FW              \\ \cline{1-5}
Baseline    & 0.927           & 0.935           & 0.959              & 0.917            \\ \hline
Incremental & 0.669           & 0.849           & 0.939              & 0.880            \\ \hhline{-----}
\end{tabular}
\caption{$R^{2}$ of different variables of two simulators. Note that $R^2 \in [0,1]$ and higher values indicate higher accuracy.}
\label{tab:goodness_of_fit}
\end{table}

\section{Simulation Experiment}

In this section, we aim to answer the following questions:
\begin{enumerate}
    \item How does our incremental NN-based simulator perform compared to an expert rule-based simulator?
    \item How effective are the control strategies optimized by the iGrow decision-making module?
\end{enumerate}

\subsection{Dataset}
There are two datasets of tomato planting involved in the simulation experiments, denoted by $D_r$ and $D_v$, respectively. 
Both datasets contain comprehensive records (control strategy, monitored sensor data and economic effectiveness) of autonomous greenhouse tomato planting process. To be specific, the data in $D_r$ are trajectories\footnote{\url{https://data.4tu.nl/articles/dataset/Autonomous_Greenhouse_Challenge_Second_Edition_2019_/12764777/2}\label{AGCSE2019_dataset}} of six independent greenhouses each controlled by a different participating team during the 2nd Autonomous Greenhouse Challenge\footnote{\url{https://www.wur.nl/en/project/autonomous-greenhouses-2nd-edition.htm}}, while  $D_v$ is composed of thousands of virtual trajectories generated by a SOTA rule-based tomato simulator (namely WUR simulator)~\cite{luo2005simulation} using multiple stochastic strategies.
Note that the two aforementioned datasets share the same calculation method of economic effectiveness\textsuperscript{\ref{AGCSE2019_dataset}}.

\subsection{Simulator evaluation}
To evaluate the performance of the incremental NN-based three-stage simulator in representing the dynamic process of complete autonomous greenhouse planting, we design the following two experiments.

\subsubsection{Accuracy on virtual trajectories.}

We consider two variants, baseline and incremental simulators, which are trained by pure virtual and hybrid trajectories, respectively.
For pure virtual trajectories, we randomly select 1000 trajectories from $D_v$, denoted by $\hat{D}_v$. For hybrid trajectories $\hat{D}_{v+r}$, we replacing 5 virtual trajectories in $\hat{D}_v$ with 5 real trajectories from $D_r$ (excluding the planting trajectory of the champion team --- \textit{Automatoes}).
Besides, both simulators are evaluated on the same test set $T_v$ consisting of 200 additional random virtual trajectories from $D_v$.
The evaluation metric is goodness-of-fit, denoted by $R^2$~\cite{mcdonald1989index}.

The $R^2$ of both simulators on different state variables are given in Table~\ref{tab:goodness_of_fit} (See Supplementary Table S1 for the descriptions of state variables).
It is easy to calculate that the overall $R^2$ of baseline and incremental simulator reach 0.911$\pm$0.009 and 0.906$\pm$0.046, respectively.
The results show that both variants of our simulator achieve accuracy comparable to that of WUR simulator, which demonstrates the feasibility and effectiveness of NN-based simulator in reducing the reliance on expert knowledge.
On the other hand, we note that the performance of incremental simulator is slightly inferior to that of incremetal simulator. 
The reason is that incremental simulator is trained on $\hat{D}_{v+r}$ containing real trajectories, whose data distribution may deviate from the test set $T_v$ composed of virtual trajectories.

\subsubsection{Accuracy on real trajectories.}
\begin{figure}[t]
    \centering
    \includegraphics[width=0.84\linewidth]{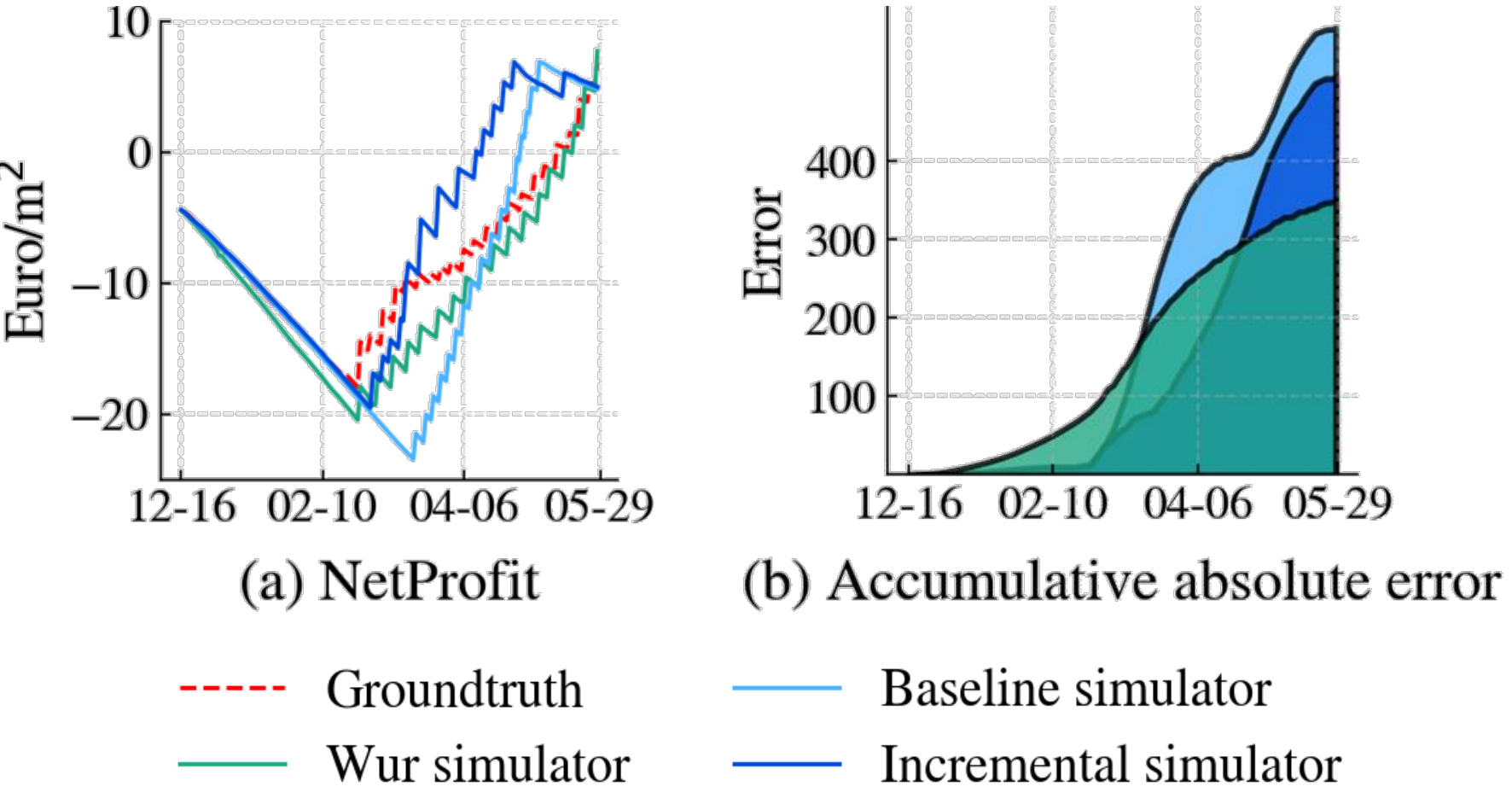}
    \caption{
    Accuracy of different simulators in the real scenario, take planting trajectory of \textit{Automatoes} as an example.
    (a) presents accumulative net profit simulated by different simulators versus ground truth;
    (b) presents accumulative absolute error of the simulated curves of net profit compared to the ground truth over the planting period.}
    \label{fig:compare_simulators}
\end{figure}

To further investigate the ability of different simulators to characterize the real planting process, we input the control strategy of champion team \textit{Automatoes} into three simulators, i.e., WUR simulator and both of our simulators.
We visualize net profit curves simulated by the three simulators versus ground truth in Figure~\ref{fig:compare_simulators}.

As shown in Figure~\ref{fig:compare_simulators}, it can be found that the gap between WUR simulator and real planting process is the smallest. 
However, WUR simulator is a rule-based simulator parametrized based on strong prior and assumptions, which makes it difficult to generalize and adapt to different real-world scenarios.
On the other hand, by comparing two variants of our simulators (see in Figure~\ref{fig:compare_simulators}(b)), we find that real data can calibrate the gap between simulation and reality.
And it can be anticipated that as the collected real data become abundant enough, incremental NN-based simulators would overtake ruled-based simulators due to stronger expressive power and fewer constraints from assumptions.

\begin{figure}[t]
    \centering
    \includegraphics[width=0.84\linewidth]{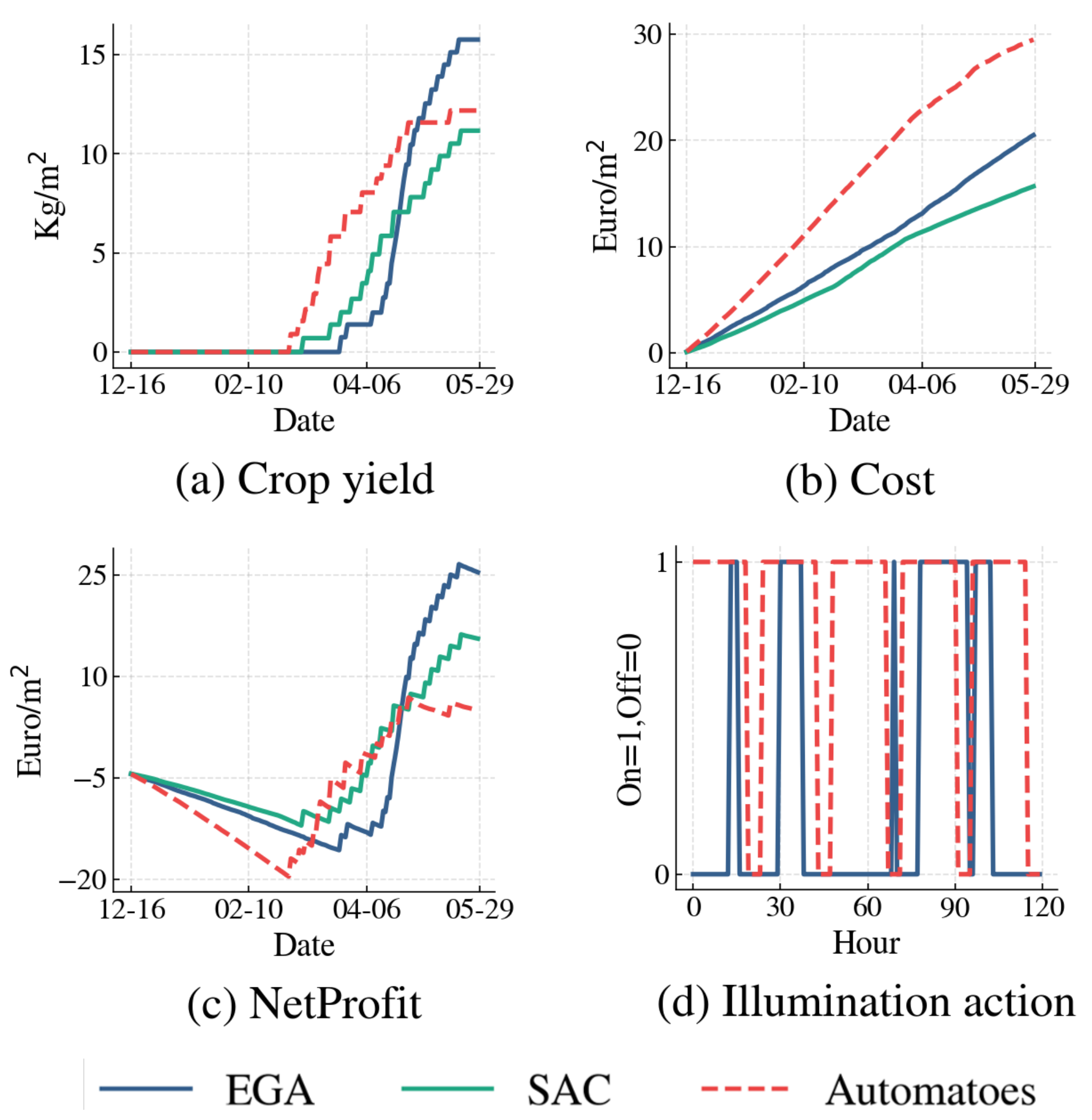}
    \caption{Performance comparison of different methods on our incremental simulator.
    (a)-(c) denote economic effectiveness simulated by different methods;
    (d) shows the 120-hour illumination action provided by EGA and \textit{Automatoes}.}
    \label{fig:compare_method}
\end{figure}

\subsection{Decision-making module evaluation}
\label{sec:comparison_methods}

In order to solve the AGC optimization problem, we propose a decision-making module.
In this section, we verify the performance of control strategies optimized by this module compared with that of planting experts.
To be specific, we use two typical optimization algorithms, SAC and EGA, to optimize the AGC strategies on the above trained incremental simulator (see parameters in Technical Appendix Section 2.2 and 2.3).
Besides, we simulate the planting strategy of \textit{Automatoes} (the champion team of the 2nd Autonomous Greenhouse Challenge), which is representative of the most advanced planting level based on the expert decision.
For the experiment to be comparable, we set the simulated outside weather to be consistent with that during the 2nd Autonomous Greenhouse Challenge.
The evaluation results are shown in Figure~~\ref{fig:compare_method},
and we can observe that:
\begin{itemize}

\item Although only EGA beats \textit{Automatoes} in yield, both optimization algorithms show significant advantages in terms of resource efficiency (see in Figure~\ref{fig:compare_method}(a), (b)).
As a result, compared to \textit{Automatoes}, the control strategies of SAC and EGA algorithms are superior at balancing yield and cost, improving net profit by 29.39\% and 285.82\%, respectively (see in Figure~\ref{fig:compare_method}(c)). 
The main reason is that optimization algorithms can provide more fine-grained control strategies than human experts, as shown in Figure~\ref{fig:compare_method}(d).
This experimental result demonstrates the potential of iGrow decision-making module in solving the AGC problem. 

\item The final net profit of SAC is inferior to that of EGA.
This is probably because SAC focuses on short-range optimization due to bootstrapping of discounted cumulative net profit by Bellman update, whereas the long-term planting period (nearly 4000 steps) poses a great challenge to the estimation accuracy.
In contrast, EGA only cares about the final net profit, by directly optimizing the full strategy of the whole planting period, thus enables a better balance between yield and cost.  

\end{itemize}

Based on the above results and analysis, we use EGA algorithm to drive the decision-making module when we deploy iGrow in real greenhouses.

\section{Real-world Experiment}

In this section, we analyze the results of 2 pilot projects deployed in real autonomous greenhouses to validate the effectiveness and superiority of iGrow in practical applications.

\subsection{Deployment overview}

We configure the iGrow supporting hardware and software facilities (decision-making, IoT and cloud-native module, etc.) in some real greenhouses, as shown in Figure~\ref{fig:framework}.
Growers/Central computers can make decisions based on the data collected by the sensors, and then send control commands to these autonomous greenhouses remotely. Next, these commands take effect via the actuators installed in greenhouses. The above process forms a closed-loop control to plant in autonomous greenhouses (see details in Supplementary Figure S2, S3 and Technical Appendix Section 3).

In terms of economic effectiveness, we refer to the calculation of the 2nd Autonomous Greenhouse Challenge, and additionally take into account the equipment depreciation (apportioning the purchase price of equipment to six years).

\subsection{Case study}

We conduct 2 pilot projects in some real autonomous greenhouses in Liaoyang City, Liaoning Province, China, with tomatoes as the experimental crop (See Supplementary Figure S1).

In each pilot project, we take a control experiment to evaluate the greenhouse control strategy, where the control groups are managed by planting experts, while the experimental groups rely on iGrow.
The 1st pilot project runs from October 2019 to March 2020, and the net profit of the experimental group (2 greenhouses) is on average about 500 \texteuro \  per greenhouse higher than that of the control group (1 greenhouse), where the area of each greenhouse is 667 $m^2$.
To further verify the statistical significance of the experimental results, we scale up the experiments in the 2nd pilot project (from March 2020 to July 2020), i.e., the control and experimental groups consist of 2 and 7 greenhouses, respectively. 

\begin{figure}[t]
    \centering
    \includegraphics[width=\linewidth]{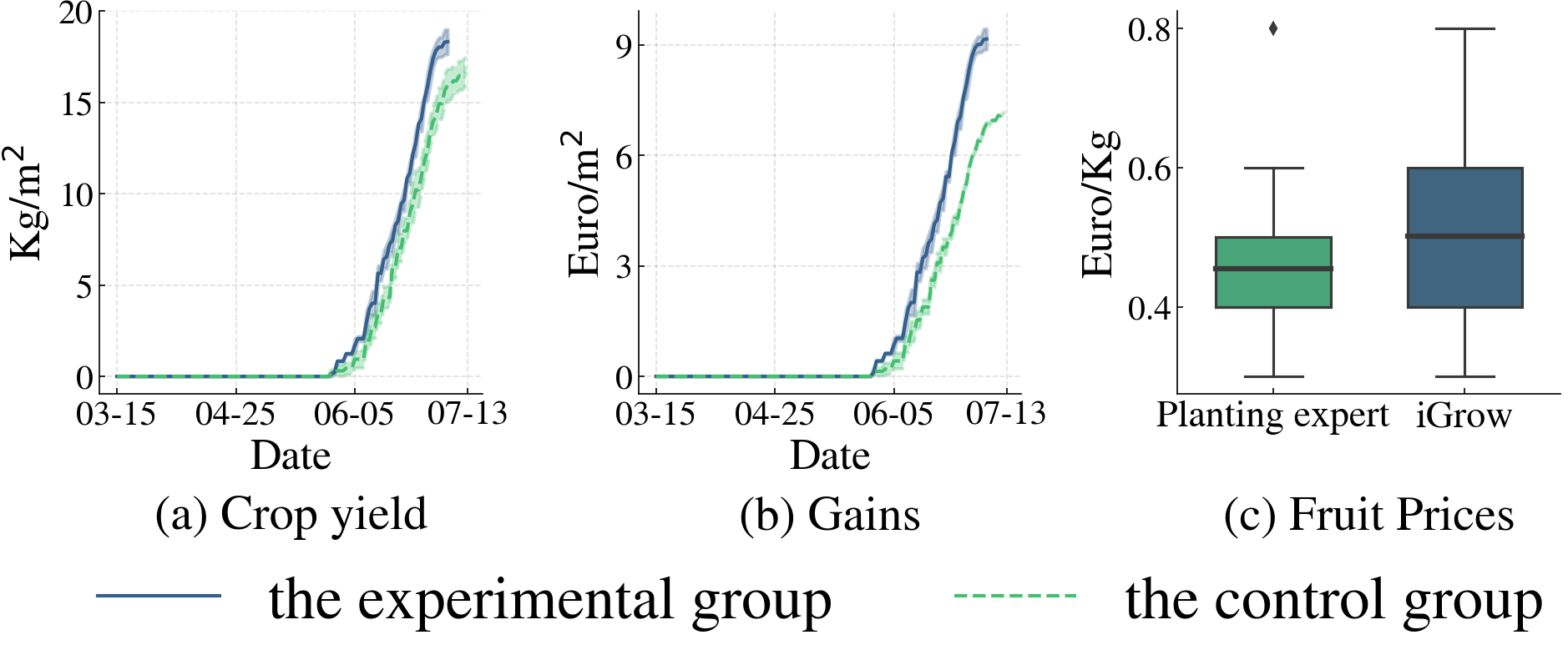}
    \caption{Comparison of the economic effectiveness of the control group (planting experts) and the experimental group (iGrow) in the 2nd pilot project.
    (a), (b) denote the accumulative harvest during the planting period.
    (c) shows the distribution of fruit prices.
    }
    \label{fig:liaoyang2_harvest}
\end{figure}

In this paper, we present the analyses of the result of the second pilot project
We use an independent sample \textit{t-test} for the economic effectiveness of the experimental group, demonstrating the superiority of iGrow compared to expert growers.
Key statistics from Figure~\ref{fig:liaoyang2_harvest} and Table ~\ref{tab:economic} are summarized below.

\begin{itemize}
    \item The experimental group on average increases yield and net profit by 10.15\% and 92.70\% compared to the control group.
    
    \item In terms of cost, although the experimental group consumes more energy, its fruits ripened faster (on average, the harvest is completed one week earlier, shown in Figure~\ref{fig:liaoyang2_harvest}(a)), resulting in significant savings in crop maintenance costs and thus lower total costs than that of the control group.
    
    \item The average unit price of fruits is higher in the experimental group, which indicates that the experimental group produces fruits of higher quality (e.g., sweetness and weight) compared to the control group.

\end{itemize}

\begin{table}[t]
\scalebox{0.7}{
\begin{tabular}{c|c|c|c|c} \hline
\makecell[c]{Economic \\ (per greenhouse)}        & \makecell[c]{Control \\ Group}    & \makecell[c]{Experimental \\ Group} & \textit{RI} & \textit{T-test} \\ \hline \hline
Energy Cost (\texteuro)            & 13.90 $\pm$ 0.23    & 50.11 $\pm$ 20.48     & -260.57\%     & 2e-2   \\ \hline
\makecell[c]{Crop Maintenance\\Cost (\texteuro) }             & 1543.96 $\pm$ 0.00     & 1457.19 $\pm$ 5.11     & +5.62\%        & 5e-6 \\ \hline
\makecell[c]{Equipment\\Depreciation (\texteuro)} & 1711.88 $\pm$ 0.00     & 1711.88 $\pm$ 0.00       & 0.00\%        & -      \\ \hline
\textbf{Total Cost} (\texteuro)             & 3269.74 $\pm$ 0.23  & 3219.18 $\pm$ 18.14   & +1.55\%        & 5e-3   \\ \hline
Fruit Prices (\texteuro /Kg)                 & 0.45 $\pm$ 0.00     & 0.49 $\pm$ 0.01       & +10.63\%       & 5e-4   \\ \hline
Crop Yield (Kg)             & 11097.60 $\pm$ 53    & 12223.90 $\pm$ 45      & +10.15\%       & 8e-3   \\ \hline
\textbf{Gains} (\texteuro)                 & 4768.30 $\pm$ 9.79  & 6106.88 $\pm$ 189.09  & +28.07\%       & 1e-4   \\ \hline
\textbf{Net Profit} (\texteuro)             & 1498.56 $\pm$ 10.03 & 2887.70 $\pm$ 183.45  & +92.70\%       & 1e-4   \\ \hline
\end{tabular}}
\begin{tablenotes}
\footnotesize{
\item \textit{RI}: relative improvement of the experimental group compared to the control group, and +/- in this column indicates improvement/degradation in performance, respectively.

\item \textit{T-test}: a value less than 0.01 indicates that the result is statistically significant.
}
\end{tablenotes}
\caption{Overall economic effectiveness comparison of the 2nd pilot project.
Note that we first account for the economic indicators for each greenhouse, and then calculate the statistical indicators (including mean and standard deviation) for the control and experimental groups, respectively.
}
\label{tab:economic}
\end{table}

Furthermore, we analyze the pair relationship among four action variables over time to understand the differences of strategies and obtain some insights that may inspire growers. Due to space limitations, the relevant visualization and analysis are described in Supplementary Figure S4 to S13.

\section{Conclusion}

In this paper, we formulate AGC as a MDP optimization problem and propose a smart agriculture solution, namely iGrow. 
The core component of iGrow is the decision-making module, which is updated by our proposed bi-level optimization algorithm.
Both simulated and pilot results demonstrate the effectiveness and superiority of iGrow.

Our solution has been verified to improve the automation level of greenhouse control and boost growing efficiency in real production environments, thus can be considered as an alternative of traditional manual control.
Moreover, the idea of bi-level optimization provides a paradigm of first building an accurate digital twin of the corresponding real-world application, and then evaluating and iterating the optimization algorithm without a real environment.

\bibliography{ref}

\section{Acknowledgments}

This work was done when Xiaoyan Cao, Yao Yao, Wanpeng Zhang, and Zhicheng An worked as interns at Tencent AI Lab.
This work was supported by the National Nature Science Foundation of China (No. 61673322) and the Natural Science Foundation of Fujian Province of China (No. 2019J01002).

\end{document}